\documentclass{llncs}
\usepackage{graphicx}
\usepackage{makecell}

\usepackage{multirow}
\usepackage{color}
\usepackage{amsmath}
\usepackage{amssymb}
\usepackage{graphicx}
\usepackage{color,soul}
\usepackage{multirow}
\usepackage[square,numbers]{natbib}
\usepackage{url}
\newcommand{\repeatthanks}{\textsuperscript{\thefootnote}}

\newcommand\N{\mathbb{N}}
\newcommand\R{\mathbb{R}}

\newcommand\D{{\cal D}}
\newcommand\EE{{\mathbb{E}}}

\DeclareMathOperator{\X}{{\cal X}}
\DeclareMathOperator{\Y}{{\cal Y}}

\providecommand{\keywords}[1]
{
  \small{\textbf{Keywords---} #1}
}

\begin{document}
\title{Do We Really Need Deep Learning Models for Time Series Forecasting?}

%
%




\author{Shereen Elsayed\thanks{Both authors contributed equally } \and
Daniela Thyssens\repeatthanks \and
Ahmed Rashed \and
Hadi Samer Jomaa \and 
Lars Schmidt-Thieme}
\institute{Department of Computer Science\\ University of Hildesheim\\ 31141 Hildesheim, Germany} 
%
%
%
\maketitle
\begin{abstract}
Time series forecasting is a crucial task in machine learning, as it has a wide range of applications including but not limited to forecasting electricity consumption, traffic, and air quality. Traditional forecasting models rely on rolling averages, vector auto-regression and auto-regressive integrated moving averages. On the other hand, deep learning and matrix factorization models have been recently proposed to tackle the same problem with more competitive performance. However, one major drawback of such models is that they tend to be overly complex in comparison to traditional techniques. In this paper, we report the results of prominent deep learning models with respect to a well-known machine learning baseline, a Gradient Boosting Regression Tree (GBRT) model. Similar to the deep neural network (DNN) models, we transform the time series forecasting task into a window-based regression problem. Furthermore, we feature-engineered the input and output structure of the GBRT model, such that, for each training window, the target values are concatenated with external features, and then flattened to form one input instance for a multi-output GBRT model. 
We conducted a comparative study on nine datasets for eight state-of-the-art deep-learning models that were presented at top-level conferences in the last years. The results demonstrate that the window-based input transformation boosts the performance of a simple GBRT model to levels that outperform all state-of-the-art DNN models evaluated in this paper.

\end{abstract}
\section{Introduction}
During the course of the past years, classical parametric (autoregressive) approaches in the area of time series forecasting have to a great extent been updated by complex deep learning-based frameworks, such as "DeepGlo" \cite{sen2019think} or "LSTNet" \cite{lai2018modeling}.
To this end, authors argue that traditional approaches may fail to capture information delivered by a mixture of long- and short-term series, and hence the argument for many deep learning techniques relies on grasping inter-temporal \textit{non-linear} dependencies in the data.
These novel deep learning-based approaches have not only supposedly shown to outperform traditional methods, such as ARIMA, and straightforward machine learning models, like GBRT, but have meanwhile spiked the expectations that time series forecasting models in the realm of machine learning need to be backed by the workings of deep learning in order to provide state-of-the-art prediction results.

However, latest since the revelation of \cite{dacrema2019we} in the field of recommender systems, it becomes evident that the accomplishments of deep learning approaches in various research segments of machine learning, need to be regularly affirmed and assessed against simple, but effective models to maintain the authenticity of the progression in the respective field of research. 
Apart from the increasing complication of time series forecasting models, another motivational argument consists of the one-sidedness of approaching time series forecasting problems with regards to deep learning-based models that are being refined in the literature, thereby limiting the diversity of existing solution approaches for a problem that exhibits one of the highest levels of diversity when applied in the real world. 
In this work, we show that with a carefully configured input handling structure, a simple, yet powerful ensemble model, such as the GBRT model \cite{friedman2001greedy}, can compete and even outperform many DNN models in the field of time series forecasting.
The assessment of a feature-engineered multi-output GBRT model is structured alongside the following two research questions:


\begin{enumerate}

\item What is the effect of carefully configuring the input and output structure of GBRT models in terms of a window-based learning framework for time series forecasting?
\item How does a simple, yet well-configured GBRT model compare to state-of-the-art deep learning time series forecasting frameworks?
\end{enumerate}
To answer these questions, we chose a two-fold experimental setup, in which we address two types of forecasting tasks: uni- and multi-variate forecasting in a systematic fashion.
The aim is to evaluate the GBRT model concerning state-of-the-art deep learning approaches that have been featured in top research conferences (NeurIPS, KDD, SDM, SIGIR, ECML, ICML, CIKM, IJCAI, ICLR).
%
\newline
The overall contributions of this research study can be summarized as follows:
\begin{itemize}
    \item \textit{GBRT}: We elevate a simple machine learning method, GBRT, to the standards of competitive DNN time series forecasting models by firstly casting it into a window-based regression framework and secondly feature-engineering the input and output structures of the model, so that it benefits most from additional context information.
    \item \textit{Comparison to naively configured baselines}: To underline the importance of the input handling for time series forecasting models, we empirically evidence why the window-based input setting for GBRT improves on the prediction performance generated by the traditionally configured models, such as ARIMA and naive GBRT implementations in the realm of time series forecasting.
    \item \textit{Competitiveness}: We study the performance of the GBRT with respect to a variety of state-of-the-art deep learning time series forecasting models and demonstrate its competitiveness on two types of time-series forecasting tasks (uni- and multi-variate).
\end{itemize}

\section{Research Design}
To motivate and explain the structure of the research study and specifically the evaluation procedure followed in the experiments, we firstly state how the baselines were selected and furthermore elaborate on the forecasting tasks chosen for the evaluation.

\subsection{Collecting Baseline Papers}

To select recently published machine learning papers for our comparison study, we systematically filter the proceedings of highly acclaimed conferences (NeurIPS, KDD, SIGIR, SDM, ECML, ICML, CIKM, IJCAI, ICLR) for the years 2016 to 2020 according to the following requirements;
\begin{itemize}
    \item \textit{Topic} - only works belonging to the field of time-series forecasting are considered.
    \item \textit{Data structure}  
          - specialised data types, such as asynchronous time series and data conceptualised as a graph were excluded.
    \item \textit{Reproducibility} - the data should be publicly available         and the source codes should be provided by the authors. In cases the source code is not available, but the experimental setup is neatly documented, we replicated the experimental results from the respective publications.
    \item \textit{Computational Feasibility} - the results presented in        the works should be reproducible in a tractable manner and computable in a             reasonable amount of time.
\end{itemize}

At this point, it is worthwhile noting that there was a considerable number of works that fit conceptually but did not comply with the above-mentioned requirements, to be reproduced. Along these lines, we initially considered, for example the approaches by \cite{qi2017mixture}, \cite{che2018hierarchical} and \cite{fan2019multi}, but were prohibited by missing source codes or not publicly available datasets. An extended list of approaches that were excluded from the experimental evaluation of the present research study can be found in Appendix A. 

\begin{table*}[!htb]
 \centering
 \small
  \caption{Dataset Statistics, where $n$ is the number of time-series, $T$ is the length of the time-series, $s$ is the sample rate, $L$ is the number of target channels, $M$ is the number of auxiliary channels (covariates), $h$ is the forecasting window size and $t'$ and $\tau$ denote the amount of training and testing time-points respectively.}
  \label{tab:1}
  \setlength{\tabcolsep}{3pt}
  \begin{tabular}{lrrrrr||rrr}
    \Xhline{2\arrayrulewidth}
    \centering
    \quad &\multicolumn{5}{c}{Data}&\multicolumn{3}{c}{Forecasting Task}\\
    \hline
    Dataset&$n$&$T\phantom{al}$&$s\phantom{aal}$&$L$&$M$&$h$&$t'\phantom{all}$&$\tau\phantom{l}$ \\
    \hline
    Electricity \cite{sen2019think} & 70& 26,136 & hourly  & 1&0 & 24 & 25,968 & 168\\
    Traffic \cite{sen2019think} & 90& 10,560 &hourly & 1&0 &24 & 10,392 & 168\\
    ElectricityV2 \cite{lim2020temporal} & 370& 6000 & hourly  & 1&0 & 24 & 5832 & 168\\
    TrafficV2 \cite{lim2020temporal} & 963& 4151 &hourly & 1&0 &24 & 3983 & 168\\
    PeMSD7(M) \cite{sen2019think}& 228 & 12,672 & 5 mins & 1&0 &9 & 11,232 & 1,440\\
    Exchange-Rate \cite{lai2018modeling} & 8 & 7,536 & daily & 1&0 &24 & 6,048 & 1,488\\
    Solar-Energy \cite{lai2018modeling} & 137 & 52,600 & 10 mins & 1&0 & 24 & 42,048 & 10512 \\
    Beijing PM2.5 \cite{du2019deep}& 1& 43,824 & hourly & 1&16 & 1, 3, 6 & 35,064 & 8,760\\
    Urban Air Quality \cite{du2019deep}& 1& 2,891,387 & hourly& 1&16 & 6 & 1,816,285 & 1,075,102\\
    SML 2010 \cite{qin2017dual}& 1 & 4,137 & per min  & 1& 26& 1 & 3,600 & 537 \\
    NASDAQ 100 \cite{qin2017dual}& 1 & 40,560 & per min & 1&81 & 1 & 37,830 & 2,730 \\
  \Xhline{2\arrayrulewidth}
\end{tabular}
\end{table*}

\subsection{Evaluation}

The evaluation of the configured GBRT model for time series forecasting is conducted on two levels; a \textit{uni}- and a \textit{multi-variate level}.
To allow for significant comparability among the selected deep learning baselines and GBRT, we assessed all models on the same pool of datasets, which are summarized in Table~\ref{tab:1} below. To this end, some datasets, specifically \textit{Electricity} and \textit{Traffic}, were initially sub-sampled for the sake of comparability (notably in Table \ref{tab:2}), but additional head-to-head comparisons concerning the affected baselines' original experimental setting are provided to validate the findings. For the experiments in Table \ref{tab:2} we therefore re-evaluated and re-tuned certain baseline models according to the changed specifications in our own experiments.
Furthermore, we adapt the evaluation metric for each baseline model separately, such that at least one of the evaluation metrics in Appendix B is used in the original evaluation of almost all considered baseline papers.

\subsubsection{Uni- and Multi-variate Forecasting Settings}
Table~\ref{tab:1} displays the statistics, as well as the respective experimental setups for the considered datasets.
The left-hand side provides information concerning the datasets itself that are used to evaluate the models, whereas the right-hand side sets forth the respective experimental specifications. The following uni-variate datasets are retrieved from \cite{sen2019think}: \textit{Traffic}, \textit{Electricity} and \textit{PeMSD7(M)}. While the \textit{Exchange Rate} and \textit{Solar Energy} datasets are collected from \cite{lai2018modeling}. We consider four multivariate datasets for the study; the \textit{Beijing PM2.5} and \textit{Urban Air Quality} dataset, both containing instances of air quality measurements from a single location ($n$=1) and additional meteorological information, in terms of $M$ covariates. The \textit{SML 2010} dataset and the \textit{Nasdaq 100} stock dataset, contain room temperature and stock market data respectively.

\section{Problem Formulation}
\textbf{Notation.} Let $\X, \Y\subseteq\R$ be sets.
For a set $\X$, let $\X^*:=\bigcup_{T\in\N}\X^T$ be finite sequences in $\X$.
For $x\in\X^T\subseteq\X^*$ ($T\in\N$), denote by $|x|:=T$ the length of $x$.
For $x$ and $y$ being vectors, we denote by $\X$ the predictor space and by $\ \Y$ the target space.
\\
\newline
\textbf{The time series forecasting problem.}
Time series forecasting, in terms of a supervised learning problem, can be formulated as follows:\\
Given a set $\X:=(\R^M\times \R^L)^*$ and a set $\Y := \R^{h \times L}$, with $M,L,h\in\N$, a sample $\D \in (\X \times \Y)^*$ from an unknown distribution $p$ and a loss function $\ell: \Y \times \Y \rightarrow \R$, find a function $\hat{y}: \X \rightarrow \Y$ called model that minimizes the expected loss:
\begin{equation}
  \label{eq:1}
    \min \phantom{a} \EE_{((x,y),y')\sim p}( \ell(y', \hat y(x,y)) )
\end{equation}
The predictors in Equation \ref{eq:1} consist of sequences of vector pairs, $(x,y)$, whereas the target sequence is denoted by $y'$. In this paper, we will look at two simple, but important special cases:
(i) the \textbf{univariate time series forecasting problem} for which there is only a single channel, $L=1$ and no additional covariates are considered, i.e. $M=0$, such that the predictors consist only of sequences of target channel vectors, $y$.
(ii) the \textbf{multivariate time series forecasting problem with a single target channel}, where the predictors consist of sequences of vector pairs, $(x,y)$, but the task is to predict only a single target channel, $L=1$.
In both cases, it is hence sufficient for the model, $\hat{y}$, to predict the target channel only:
    $\hat y: (\R^{h\times L})^*\rightarrow\R$.
\section{Feature-engineered window-based GBRT}
\begin{figure*}
\centering
    \includegraphics[clip,trim=0.7cm 0.0cm 0.0cm 0.0cm,width=1.0\textwidth]{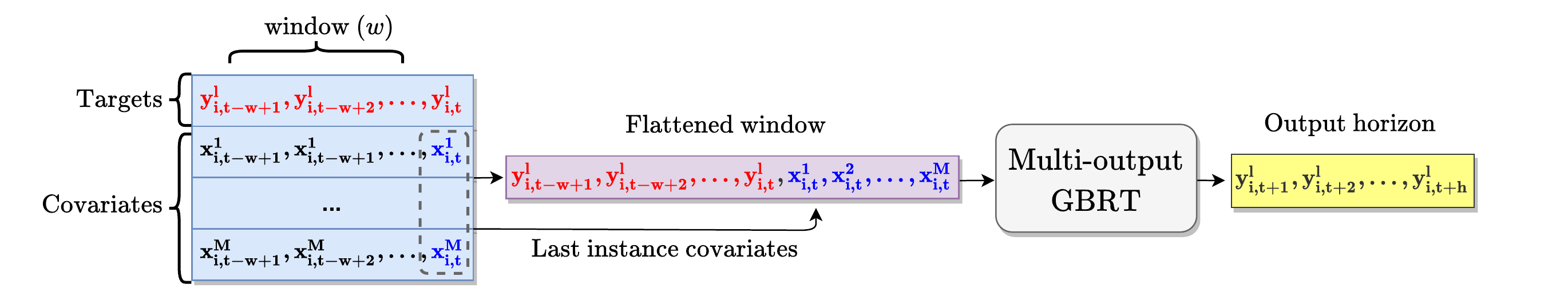}
    \caption{Reconfiguration of a single window of input instances. $Y_{i,t-w+1}^{l},\ldots,Y_{i,t}^{l}$ are the target instances in the window, whereas $X_{i,t-w+1}^{1},\ldots,X_{i,t}^{M}$ are the (external) features in the window for a particular time series $i \in n$.}
    \label{fig:2}
\end{figure*}

The investigated GBRT model \cite{friedman2001greedy}, specifically the XGBoost implementation \cite{chen2016xgboost} thereof, brings forth the benefits that it is easy to apply and particularly successful on structured data.
But when it comes to the naive implementation in time series forecasting \cite{papadopoulos2015short,chen2018neucast}, GBRT models lose a great part of their flexibility, because they are not casted into window-based regression problems, but instead configured, such that they are fitted on the majority of the time series as a complete and consecutive sequence of data points to predict the subsequent and remaining testing part of the time series.
Unlike this naive way of input handling, we simulate the input processing behavior used in successful time series forecasting models by re-configuring the time series into windowed inputs and train on those multiple training instances (windows) instead, for which we define a tunable window size, $w \in \N$. This window-based input setting for GBRT models is illustrated in Figure \ref{fig:2}; the first step is to reshape the typical 2D training instances (time series input windows) into GBRT-suitable 1D-shaped vectors (flattened windows) using a transformation function $\phi: \mathbb{R}^{w  \times (L+M)}  \rightarrow \mathbb{R}^{(w + M)}$. This function concatenates the target values $y_i$ from all $w$ instances and then appends the covariate-vector of the \textit{last time-point instance} $t$ in the input window $w$, denoted as $X_{i,t-w+1}^{1},\ldots,X_{i,t}^{M}$ in Figure \ref{fig:2}.
After this reformulation, we obtain the input sequence vector pairs in Equation \ref{eq:1} for a single time series $i \in n$ and single target channel $l\in L$ as $(x,y)=(Y_{i,t-w+1}^{l},\ldots,Y_{i,t}^{l};X_{i,t}^{1},\ldots,X_{i,t}^{M})$. Regarding the choice of appending only the last time-point instance's covariates, we refer to the ablation study in section 5.3. The next and final step consists of passing these input vectors to the multi-output GBRT to predict the future target horizon $y'=Y_{i,t+1}^{l},\ldots,Y_{i,t+h}^{l}$ for every instance, where $h \in \N$ denotes the amount of time steps to be predicted in the future.
The multi-output characteristic of our GBRT configuration is not natively supported by GBRT implementations, but can be instantiated through the use of problem transformation methods, such as the single-target method \cite{borchani2015survey}. 
In this case, we chose a multi-output wrapper transforming the multi-output regression problem into several single-target problems. This method entails the simple strategy of extending the number of regressors to the size of the prediction horizon, where a single regressor, and hence a single loss function (\ref{eq:1}), is introduced for each prediction step in the forecasting horizon. The final target prediction is then calculated using the sum of all tree model estimators.
This single-target setting automatically entails the drawback that the target variables in the prediction horizon are forecasted independently and that the model cannot benefit from potential relationships between them. Which is exactly why the emphasise lies on the window-based input setting for the GBRT that not only transforms the forecasting problem into a regression task, but more importantly allows the model to capture the auto-correlation effect in the target variable and therefore compensates the initial drawback of independent multi-output forecasting.
The described window-based GBRT input setting drastically lifts its forecasting performance, as the GBRT model is now capable of grasping the underlying time series structure of the data and can now be considered an adequate machine learning baseline for advanced DNN time series forecasting models.
The above-mentioned naively configured GBRT model $\hat y_{naive}$ on the other hand, is a simple point-wise regression model that takes as input the concurrent covariates of the time point $X_{i,j}^{1},\ldots,X_{i,j}^{M}$ and predicts a single target value $Y_{i,j}$ for the same time point such that the training loss;
\begin{equation}
 \label{eq:2}
 \small
     \ell(\hat y; Y):= \frac{1}{n} \sum_{i=1}^n  \frac{1}{t'} \sum_{j=1}^{t'} (Y_{i,j}- \hat y_{naive}(X_{i,j}^{1},\ldots,X_{i,j}^M))^2 
\end{equation}
is minimal.


\section{Experiments and Results}

The experiments and results in this section aim to evaluate a range of acclaimed DNN approaches on a reconfigured GBRT baseline for time series forecasting. We start by introducing these popular DNN models in section 5.1 and thereafter consider the uni- and multi-variate forecasting setting in separate subsections. Each subsection documents the results and findings for the respective tasks at hand. Concerning the experimental protocol, it is to note that the documented results in the tables are retrieved from a final run including the validation data portion as training data. Our codes are available and accessible through a github repository\footnote{https://github.com/Daniela-Shereen/GBRT-for-TSF}.

\subsection{DNN Time Series Forecasting Approaches}
The following prominent deep learning based models are considered for the assessment in this study:
\begin{enumerate}
    \item The Temporal Regularized Matrix Factorization (TRMF) model \cite{yu2016temporal} is a highly scalable matrix factorization based approach, due to its ability to model global structures in the data. Being one of the earlier approaches in this study, the model is restricted to capture linear dependencies in time series data, but has nevertheless shown highly competitive results. 
    \item The Long- and Short-term Time-series Network (LSTNet) \cite{lai2018modeling} emphasises both, local multivariate patterns, modeled by a convolutional layer and long-term dependencies, captured by a recurrent network structure. Originally, LSTNet \cite{lai2018modeling} featured two versions; "LSTNet-Skip" and "LSTNet-Attn". Since "LSTNet-Attn" is not reproducible\footnote{https://github.com/laiguokun/LSTNet/issues/11}, "LSTNet-Skip" is evaluated in the subsequent experiments.
    \item The Dual-Stage Attention-Based RNN (DARNN) \cite{qin2017dual} firstly passes the model inputs through an input attention mechanism and subsequently employs an encoder-decoder model equipped with an additional temporal attention mechanism. The model is originally evaluated on two multi-variate datasets, but is also featured in the univariate assessment in section 5.2, due to its straightforward applicability to univariate datasets.
    \item The Deep Global Local Forecaster (DeepGlo) \cite{sen2019think} is based on a global matrix factorization structure that is regularized by a temporal convolutional network. The model incorporates additional channels derived from the date and timestamp and is originally assessed on univariate datasets.
    \item The Temporal Fusion Transformer (TFT) model \cite{lim2020temporal} is the most recent DNN approach featured in this study. The powerful framework combines recurrent layers for local processing with the transformer-typical self-attention layers that capture long-term dependencies in the data. Not only can the model dynamically attend to relevant features during the learning process, it additionally suppresses those qualified as irrelevant through gating mechanisms.
    \item The DeepAR model \cite{salinas2019deepar} is an auto-regressive probabilistic RNN model that estimates parametric distributions from time series with the help of additional time- and categorical covariates. As the open source implementation of DeepAR (GluonTS \cite{alexandrov2019gluonts}) claims itself to be only "similar" to the architecture described in \cite{salinas2019deepar}, we preferrably compare head to head with published normalized deviation results (referred to as WAPE in the subsequent result tables)  instead of re-implementing it from scratch.
    \item The Deep State Space Model (DeepState) \cite{rangapuram2018deep} is a probabilistic generative model that learns to parametrize a linear state space model using RNNs. Similar to DeepAR, the open source implementation is technically available through GluonTS \cite{alexandrov2019gluonts}, but in this case the hyperparameters for reproducing the published results were not clearly stated in \cite{rangapuram2018deep}, which encouraged us to also engage in a head to head comparison with regards to the published normalized deviation results.
    \item The Deep Air Quality Forecasting Framework (DAQFF) \cite{du2019deep} consists of a two-staged feature representation; The data is passed through three 1D convolutional layers, followed by two bi-directional LSTM layers and a subsequent linear layer for prediction. As is deducible from the model name, this framework is explicitly constructed to predict air quality and is thus assessed on the respective multivariate datasets listed in Table \ref{tab:1}. 
    \end{enumerate}
On another note, to address research question 2 and highlight the importance of casting time series forecasting problems as rolling window-based regression problems, two further baseline models are included; a naively configured GBRT baseline (GBRT(Naive)) and a simple ARIMA model. Opposed to the window-based GBRT (GBRT(W-b)), these naively configured implementations consists of firstly fitting the respective model on the complete and consecutive training portion of the data, before evaluating it on the remaining testing data as is.

\subsection{Univariate Datasets}
Univariate time series forecasting aims at predicting a single target variable in the future, based on the historical input of this target variable only. At this point it is important to mention that we consider the existence of multiple ($n$) independent single-targeted time series of the same variable not as a characteristic of multi-variate time series forecasting, but rather as multiple, independent uni-variate time series. We furthermore allow for the construction of simple time-covariates that are, for most cases, extracted from the timestamp information. 
For the datasets in this subsection, it conviniently just so happened that the lookup window size used is equivalent to the forecasting window size, that can be deduced from Table \ref{tab:1}.




\subsubsection{Results for Uni-variate Time Series}
The results in table \ref{tab:2} summarize the forecasting performances concerning the uni-variate time series forecasting datasets without using simple covariates as predictors. Overall the results indicate strong competitiveness on behalf of the window-based GBRT with the sole exception of traffic forecasting. Traditionally configured forecasting models, such as ARIMA and GBRT(Naive), are, on the other hand, expectably outperformed by far. This finding emphasises the relevance of carefully configuring and adapting machine learning baseline to the given problem. While for this uni-variate setting, no covariates were considered, the performance gains in GBRT(W-b)  can exclusively be attributed to the rolling forecasting formulation for GBRT. 
\begin{table*}[!htb]
\caption{Experimental Results for Univariate Datasets without covariates (bold represents the best result and underlined represents second best)}
\label{tab:2}
\begin{center}
\begin{tabular}{cccccccc}
    \Xhline{2\arrayrulewidth}
    \multicolumn{8}{c}{Without covariates} \\
    \hline
    Dataset& & LSTNet & TRMF & DARNN & GBRT\scriptsize{(Naive)}&ARIMA&GBRT\scriptsize{(W-b)} \\
    \hline
    \multirow{3}{*}{Electricity}&RMSE&1095.309&\underline{136.400}&404.056&523.829&181.210& \textbf{125.626}\\
    &WAPE&0.997&\textbf{0.095}&0.343&0.878&0.310&\underline{0.099}\\
    &MAE&474.845&\textbf{53.250}&194.449&490.732&154.390&\underline{55.495}\\
    \hline
    \multirow{3}{*}{Traffic}&RMSE&0.042&\underline{0.023}&\textbf{0.015}&0.056&0.044&0.046\\
    &WAPE&\textbf{0.102}&0.161&\underline{0.132}&0.777&0.594&0.568\\
    &MAE&0.014&\underline{0.009}&\textbf{0.007}&0.043&0.032&0.030\\
    \hline
    \multirow{3}{*}{PeMSD7}&RMSE&55.405 &\textbf{5.462}  &5.983&12.482 &15.357&\underline{5.613} \\
    &WAPE&0.981&\underline{0.057}& 0.060& 0.170&0.183&\textbf{0.051} \\
    &MAE&53.336&\underline{3.329}&3.526&9.604&10.304&\textbf{3.002}\\
    \hline
    \multirow{3}{*}{Exchange-Rate}&RMSE&\underline{0.018}&\underline{0.018}&0.025& 0.081&0.123&\textbf{0.017}\\
    &WAPE&0.017&\underline{0.015}&0.022&0.456&0.170&\textbf{0.013}\\
    &MAE&0.013&\underline{0.011}& 0.016 &0.068&0.101&\textbf{0.010} \\
     \Xhline{2\arrayrulewidth}
\end{tabular}
\end{center}
\end{table*}
\newline
Whereas for electricity forecasting, the window-based GBRT shows the best RMSE performance across all models with a respectable margin, its performance concerning WAPE and MAE is solely outperformed by TRMF introduced in 2016.
The attention-based DARNN model exhibits worse performance, but has originally been evaluated in a multi-variate setting on stock market and indoor temperature data. Unlike LSTNet, which has originally been evaluated in a uni-variate setting, but had to be re-implemented for all datasets in Table \ref{tab:2}, due to the different evaluation metrics in placed. 
Regarding the exchange rate prediction task, LSTNet (being re-implemented with $w =24$) and TMRF show comparably strong results, but are nevertheless outperformed by the window-based GBRT baseline. Due to the unfavourable performance results on behalf of LSTNet in Table \ref{tab:2}, affirming results are shown in Table \ref{tab:4} with respect to its originally used metrics and original experimental setup.
Without considering time-predictors, the results for traffic forecasting are mixed, such that the best results for the hourly \textit{Traffic} dataset are achieved by DARNN and LSTNet, while for the \textit{PeMSD7} dataset, the window-based GBRT baseline outperforms the DNN models on two out of three metrics. 
The inclusion of time covariates however, boosts GBRT's performance considerably (Table \ref{tab:3}), such that, also for traffic forecasting, all DNN approaches, including DeepGlo \cite{sen2019think} and a popular spatio-temporal traffic forecasting model (STGCN) \cite{yu2017spatio}, which achieved an RMSE of 6.77 on PeMSD7, are outperformed by the reconfigured GBRT baseline.
\begin{table*}[!htb]
\caption{Experimental Results for Univariate Datasets with covariates}
\label{tab:3}
\begin{center}
\begin{tabular}{ccccc}
    \Xhline{2\arrayrulewidth}
    \multicolumn{5}{c}{With time-covariates} \\
    \hline
    Dataset& & DeepGlo & GBRT\scriptsize{(Naive)}&GBRT\scriptsize{(W-b)} \\
    \hline
    \multirow{3}{*}{Electricity}&RMSE& \underline{141.285}& 175.402&\textbf{119.051}\\
    &WAPE&\underline{0.094}& 0.288&\textbf{0.089}\\
    &MAE& \underline{53.036}&143.463&\textbf{50.150}\\
    \hline
    \multirow{3}{*}{Traffic}&RMSE& \underline{0.026}& 0.038&\textbf{0.014}\\
    &WAPE& \underline{0.239}&0.495&\textbf{0.112}\\
    &MAE& \underline{0.013}&0.027&\textbf{0.006}\\
    \hline
    \multirow{3}{*}{PeMSD7}&RMSE& \underline{6.490*}&8.238&\textbf{5.194} \\
    &WAPE&\underline{0.070}&0.100&\textbf{0.048}\\
    &MAE&\underline{3.530*}&5.714&\textbf{2.811}\\
    \hline
    \multirow{3}{*}{Exchange-Rate}&RMSE& \underline{0.038}&0.079&\textbf{0.016}\\
    &WAPE&\underline{0.038}&0.450&\textbf{0.013}\\
    &MAE& \underline{0.029}&0.066&\textbf{0.010}\\
     \Xhline{2\arrayrulewidth}
\end{tabular}
\\ \small (*) Results reported from the original paper.
\end{center}
\end{table*}
\newline
Overall, windowing inputs and adding simple time-covariates to gradient-boosted tree models demonstrates a convincing generalization performance across the various univariate time series datasets in Table \ref{tab:2} and \ref{tab:3}. To further affirm this finding and mitigate any disadvantages for the DNN approaches that may have been caused by different evaluation metrics or subsampled datasets, the subsequent subsections comprise head to head experiments for an evaluation on published performance results.

\subsubsection{Complementary Comparison against LSTNet}
In this section, we assess LSTNet on the additional \textit{Solar-Energy} dataset showcased together with the \textit{Exchange-Rate} dataset in the original publication \cite{lai2018modeling}. 
Table \ref{tab:4} displays the results for GBRT(W-b) including time covariates and a forecasting window of $h=24$ evaluated on Root Relative Squared Error (RSE) and Empirical Correlation Coefficient (Corr). 
\begin{table}[!htb]
\centering
  \caption{LSTNet vs window-based GBRT}
  \label{tab:4}
  \addtolength{\tabcolsep}{2.0pt}
  \begin{tabular}{ccllcll}
    \Xhline{2\arrayrulewidth}
     Model &\multicolumn{2}{c}{Solar-Energy}&\multicolumn{2}{c}{Exchange-Rate}\\
    &RSE&Corr&RSE&Corr\\
    \hline
    LSTNet\text{*} \cite{lai2018modeling}&0.464&0.887&0.044&0.935\\
    GBRT\scriptsize{(W-b)}&\textbf{0.455} &\textbf{0.896} &\textbf{0.037}&\textbf{0.999}\\
  \Xhline{2\arrayrulewidth}
\end{tabular}
\\ \small (*) Results reported from the original paper.
\end{table}
These complementary results reinforce the above finding that strong, deep learning-backed frameworks, such as LSTNet, can be (consistently) outperformed by a well-configured GBRT model. 

\subsubsection{Comparison against Probabilistic and Tranformer-based Models}
Lastly, we want to affirm the above finding for univariate datasets also with respect to probabilistic models, such as DeepAR \cite{salinas2019deepar} and DeepState \cite{rangapuram2018deep}, and a Transformer-based model (TFT) \cite{lim2020temporal}. In order to compare head to head with published results, we apply the experimental setting followed in \cite{lim2020temporal} regarding using different versions of the \textit{ElectricityV2} and \textit{TrafficV2} datasets; specifically for \textit{ElectricityV2} has $n=370$ available series, but the time series length is $T=6000$, while the \textit{TrafficV2} dataset consisting of 963 series time series length around $T=4000$. The testing period, described in Table \ref{tab:1} (seven days) remains the same and simplistic timestamp-extracted covariates are used in all models.
The parameters for the window-based GBRT on the \textit{TrafficV2} dataset are the same as the ones used for the sub-sampled dataset, whereas for \textit{ElectricityV2}, the parameters had to be tuned separately.\\
\begin{table}[!htb]
\centering
  \caption{DeepAR, DeepState and TFT vs window-based GBRT. The results are given in terms of the normalized deviations (WAPE).}
  \label{tab:5}
\begin{tabular}{lccccc}
    \Xhline{2\arrayrulewidth}
     Dataset &\multicolumn{4}{c}{Model}\\
    &DeepAR\text{*} \cite{salinas2019deepar}&DeepState\text{*} \cite{rangapuram2018deep}&TFT\text{*} \cite{lim2020temporal}&GBRT\scriptsize{(W-b)}\\
    \hline
    ElectricityV2 &0.070&0.083&\textbf{0.055}&\underline{0.067}\\
    TrafficV2 &0.170&0.167&\textbf{0.095} &\underline{0.148}\\
  \Xhline{2\arrayrulewidth}
\end{tabular}
\\ \small (*) Results reported from \cite{lim2020temporal}
\end{table}
\newline
The results in Table \ref{tab:5} above underline the competitiveness of the rolling forecast-configured GBRT, but also show that considerably stronger transformer-based models, such as the TFT \cite{lim2020temporal}, rightfully surpass the boosted regression tree performance.
Nevertheless, as an exception, the TFT makes up the only DNN model that consistently outperforms GBRT in this study, while probabilistic models like DeepAR and DeepState are outperformed on these uni-variate datasets.\\
\newline
A major finding of the results in this subsection was that even simplistic covariates, mainly extracted from the timestamp, elevated the performance of a GBRT baseline extensively. In the next subsection, these covariates are extended to consist of expressive variables featured in the dataset.

\subsection{Multivariate Datasets}
The considered multi-variate time series forecasting setting describes the case where the data for more than one feature is natively provided in a dataset, however, only one single target variable needs to be forecasted. 
In this case, we are given external features $X_{i,t-w}^{1},\ldots,X_{i,t}^{M}$ that are more expressive than simplistic time-predictors extracted from the timestamp.
Concerning the features, two alternative ways exist to look at multi-variate time series forecasting in the literature, such as in \cite{qin2017dual} and \cite{du2019deep} for example, we decided to stick to the terminology in \cite{du2019deep} to differentiate between having a single target with multiple conditioning covariates and multi-time series instances.\newline
The following subsections refer to two different multi-variate forecasting tasks and their respective experimental settings; 
Firstly, we revisit the attention-based DARNN model from section 5.2 in order to evaluate its performance against the GBRT baseline in a head to head fashion. 
In a second step, we assess a DNN framework, namely DAQFF, against GBRT on the air quality prediction task and implicitly engage also in a comparison of simpler neural network structures, such as CNNs and LSTMs for this task.

\subsubsection{Comparison against DARNN with Covariates}
For this head to head comparison, the multi-variate forecasting task in \cite{qin2017dual} is to predict target values, room temperature (\textit{SML 2010}) and stock price (\textit{NASDAQ100}) respectively, one step ahead, given various predictive features and a lookup window size of 10 data points, which has also been proven to be the best value for DARNN.
\begin{table}
\centering
\small
  \caption{DARNN vs window-based GBRT}
  \label{tab:6}
   \addtolength{\tabcolsep}{1.5pt}
   \begin{tabular}{ccccccc}
    \Xhline{2\arrayrulewidth}
     Model &\multicolumn{3}{c}{SML 2010} &\multicolumn{3}{c}{NASDAQ100}\\
    &RMSE&MAPE&MAE&RMSE&MAPE&MAE\\
    \hline
    ARIMA \text{*} \cite{asteriou2011arima} &0.0265& 0.0929&0.0195 &1.4500&0.0184&0.9100   \\
    DARNN \text{*} \cite{qin2017dual}&0.0197 &0.0714&0.0150&0.3100&\textbf{0.0043} &0.2100\\
    GBRT\scriptsize{(W-b)}&\textbf{0.0168}&\textbf{0.0615}&\textbf{0.0134}&\textbf{0.0784}&0.0257&\textbf{0.0633}\\
    \hline
  \Xhline{2\arrayrulewidth}
 \end{tabular}
\\ \small (*) Results reported from \cite{qin2017dual}.
\end{table}
\newline
The results in Table \ref{tab:6} support the findings from the previous section also in this multi-variate case and show that even attention-backed DNN frameworks specifically conceptualised for multi-variate forecasting can be outperformed by a simple, well-configured GBRT baseline. 
On another note, given that the only non-DNN baseline in the evaluation protocol in \cite{qin2017dual} was ARIMA, highlights furthermore the one-sidedness of machine learning forecasting models for the field of time series forecasting. Thus, generally, care has not only to be taken when configuring presumably less powerful machine learning baselines, but also when creating the pool of baselines for evaluation.

\subsubsection{Comparison against DAQFF}
As a last comparative (head to head) experiment in this study, we assess a fully-fletched DNN model, namely Deep Air Quality Forecasting Framework, that has explicitly been constructed for the air quality forecasting task on the reconfigured GBRT baseline.
The original results concerning DAQFF \cite{du2019deep} were not reproducible, since the source codes were not available, but the data was nevertheless accessible.The original, well-documented, data pre-processing scheme and experimental setup was adopted, such that the forecasting window size was chosen to be 6 hours and the lookup window size was set to 1 hour for both datasets.
\begin{table}[!htb]
\centering
  \caption{Naive and window-based GBRT vs DAQFF on the Beijing PM2.5 and Urban Air Quality datasets for lookup window size 1 and forecasting window size 6.}
  \label{tab:7}
  \begin{tabular}{llccccccc}
    \Xhline{2\arrayrulewidth}
    Dataset&\multicolumn{8}{c}{Model}\\
    &&LSTM\text{*}&GRU\text{*} &RNN\text{*}&CNN\text{*}&DAQFF\text{*}&GBRT\scriptsize{(Naive)}&GBRT\scriptsize{(W-b)}\\
    \hline
    \multirow{2}{*}{PM2.5}&RMSE& 57.49& 52.61&57.38&52.85&\underline{43.49}&83.08&\textbf{42.37}\\
    &MAE& 44.12&38.99&44.69& 39.68&\underline{27.53}&56.37&\textbf{25.87}\\
    \hline
    \multirow{2}{*}{Air Quality}&RMSE& 58.25& 60.76&60.71&53.38&46.49&\textbf{40.50}&\underline{40.55}\\
    &MAE& 44.28&45.53&46.16& 38.21&25.01&\underline{24.30}&\textbf{22.34}\\
    
    \hline
  \Xhline{2\arrayrulewidth}
\end{tabular}
  \\ \small (*) Results reported from \cite{du2019deep}. 
\end{table}
Table \ref{tab:7} shows that even DNN models that are specifically designed for a certain forecasting task, air quality prediction in this case, and therefore are assumed to work particularly well concerning that task, are not meeting the expectations. Instead, the DAQFF is performing worse than a simple window-based, feature-engineered gradient boosted regression tree model. In this experiment, it is to note that even a GBRT model used in the traditional applicational forecasting sense delivers better results on the Air Quality Dataset. Additional experiments regarding DAQFF that support the general finding if this study are documented in Appendix C and have been omitted here due to space constraints.


\subsection{Ablation Study}
This section presents supporting results for the feature inclusion scheme, elaborated on in section 4, such that including only the last time step's covariates in the flattened GBRT input window is sufficient to achieve competitive results. 
\begin{table}[!ht]
\centering
  \caption{Performance comparison between using only the covariates of last windowed instance against using covariates of all the windowed instances.}
  \label{tab:8}
  \setlength{\tabcolsep}{4pt}
   \begin{tabular}{lrrrr}
    \Xhline{2\arrayrulewidth}
     Model &\multicolumn{2}{c}{ All instances}&\multicolumn{2}{c}{ Last instance}\\
    &RMSE&MAE&RMSE&MAE\\
    \hline
    SML2010&0.016 &0.013 &0.016 &0.013 \\
    NASDAQ&0.096 &0.078 &\textbf{0.078} &\textbf{0.063 }\\
    Exchange-Rate&0.016 & 0.010&0.016 & 0.010\\
    Beijing PM2.5&\textbf{33.310} &\textbf{19.150} &33.580 &19.250  \\
  \Xhline{2\arrayrulewidth}
 \end{tabular} 
\end{table}
In this regard, we assess both configurations of the window-based GBRT on selected datasets from Table \ref{tab:1} and show the difference in performance. 
The experimental setup concerning the datasets is the same as in the respective subsections above, except for \textit{PM2.5}, where the lookup window size and the prediction window size is set to 6 and 3 respectively.
\newline
The results in Table \ref{tab:8} demonstrate that considering only the last instance's auxiliary features has barely caused an information loss, such that a lot of computational memory and power can be saved by applying the "Last instance" scheme.


\section{Conclusion}


In this study, we have investigated and reproduced a number of recent deep learning frameworks for time series forecasting and compared them to a rolling forecast GBRT on various datasets. 
The experimental results evidence that a conceptually simpler model, like the GBRT, can compete and sometimes outperform state-of-the-art DNN models by efficiently feature-engineering the input and output structures of the GBRT.
On a broader scope, the findings suggest that simpler machine learning baselines should not be dismissed and instead configured with more care to ensure the authenticity of the progression in field of time series forecasting.
For future work, these results incentivise the application of this window-based input setting for other simpler machine learning models, such as the multi-layer perceptron and support vector machines.

\bibliography{sample-base}



\end{document}